\documentclass[preprint, a4paper, 12pt]{elsarticle}
\usepackage{hyperref}
\usepackage{booktabs}
\usepackage{multirow}
\usepackage[table,xcdraw]{xcolor}
\usepackage{amsmath}
\usepackage{bbold}
\usepackage{subfig}
\usepackage{amssymb}
\usepackage{float}

\journal{Ecological Informatics}
\begin{document}
	\begin{frontmatter}
		\title{Automated Feature-Specific Tree Species Identification from Natural Images using Deep Semi-Supervised Learning}
		\author[addr1]{Dewald Homan\corref{cor1}}
		\ead{dewald.homan@gmail.com}
		\author[addr1]{Johan A. du Preez}
		\ead{dupreez@sun.ac.za}
		\cortext[cor1]{Corresponding author}
		\address[addr1]{Faculty of Engineering, Stellenbosch University, Stellenbosch, 7602, South Africa}
		
		\begin{abstract}
			 Prior work on plant species classification predominantly focuses on building models from isolated plant attributes. Hence, there is a need for tools that can assist in species identification in the natural world. We present a novel and robust two-fold approach capable of identifying trees in a real-world natural setting. Further, we leverage unlabelled data through deep semi-supervised learning and demonstrate superior performance to supervised learning. Our single-GPU implementation for feature recognition uses minimal annotated data and achieves accuracies of 93.96\% and 93.11\% for leaves and bark, respectively. Further, we extract feature-specific datasets of 50 species by employing this technique. Finally, our semi-supervised species classification method attains 94.04\% top-5 accuracy for leaves and 83.04\% top-5 accuracy for bark.
			
		\end{abstract}
	\end{frontmatter}
	\pagebreak	
	
	\section{Introduction}	
		Species and expert human species recognition skills are vanishing concurrently \citep{pimm2014biodiversity, waldchen2018automated}. Further, species identification is fundamental to ecological studies, and trees form the foundation for forest ecosystems. Therefore, improved automated tree species classification can greatly benefit biodiversity conservation.
		
		Current species classification methods largely steer clear of images containing partial, deformed, compounded or overlapped plant features \citep{waldchen2018automated}. Consequently, suitable databases for tree species classification using natural images - images of complex, natural scenes \citep{martin2001database} - are scarce. Even so, iNaturalist \citep{inaturalist} boasts an expanding collection of global biodiversity observations, prompting the compilation of species classification datasets \citep{Horn_2018_CVPR}. Accordingly, we compile a tree species dataset using iNaturalist. Our dataset is undoubtedly more challenging than existing plant classification datasets consisting of images with single isolated tree attributes \citep{carpentier2018tree, wu2007leaf}.

		In order to assist the processing of collected taxonomic data, neural networks were first proposed for species identification nearly 30 years ago \citep{simpson1992biological, morris1992identification}. Recently, automated species classification and, specifically, plant identification has seen deep learning become increasingly popular. In particular, convolutional neural networks (CNNs) provide superior classification performance compared to conventional computer vision and machine learning techniques \citep{waldchen2018automated}.	
		
		Nevertheless, deep supervised image classification methods traditionally require a substantial number of labelled images to reach desirable results. Accordingly, it is customary for CNN-based plant recognition methods \citep{pawara2017comparing, vsulc2017fine} to use a pre-trained deep neural network model and apply supervised learning to fine-tune it to the desired dataset. However, intermissions in the publication of sufficiently large datasets and leading deep learning models may temporarily halt innovation \citep{bourlard1996towards}.
		
		In contrast, semi-supervised learning (SSL) can exploit the abundance of unlabelled data in label-scarce scenarios \citep{pise2008survey}. Regrettably, species classification pipelines seldomly employ SSL. Remote sensing applications focusing on tree crown detection incorporate SSL \citep{dalponte2015semi, weinstein2019individual}, but not for deep fine-grained species classification. 
		
		Therefore, we propose a novel deep SSL method for tree species identification. Our method demonstrates the value of SSL for natural image data with proven classification performance.
		
		We present the remainder of the contents as follows: Section~\ref{MatAndMethods} describes the compilation of our dataset and introduces deep semi-supervised methods. The results and discussion of experiments follow in Section~\ref{Results}, and we include relevant conclusions in Section~\ref{Conclusions}.	
						
	\section{Materials and Methods} \label{MatAndMethods}

		\subsection{Dataset compilation} \label{dataset}
		We introduce a natural image dataset equivalent to the 185 tree species included in \citep{kumar2012leafsnap}. Firstly, we interrogate the iNaturalist database for the species included in the \textit{Trees of Northeastern United States}\footnote{Available at https://www.inaturalist.org/projects/trees-of-northeastern-us} project and collate the images according to species. Observation images on iNaturalist are categorized as either \textit{casual}, \textit{need ID} or \textit{research grade} according to certain labelling conditions. We compile two independent datasets, research-grade and need-ID since the project has only 53\% research-grade observations.
		
		To limit our datasets to 50 species, we remove the needle-bearing trees and select the top-50 species ranked according to the number of species observations. The total number of images included in the research-grade dataset is 126 344, while the need-ID dataset contains 46 008 images. The remainder of Section~\ref{MatAndMethods} describes work done using the former dataset unless otherwise stated.  
		
\begin{figure}[ht!]
			\includegraphics[width=\linewidth]{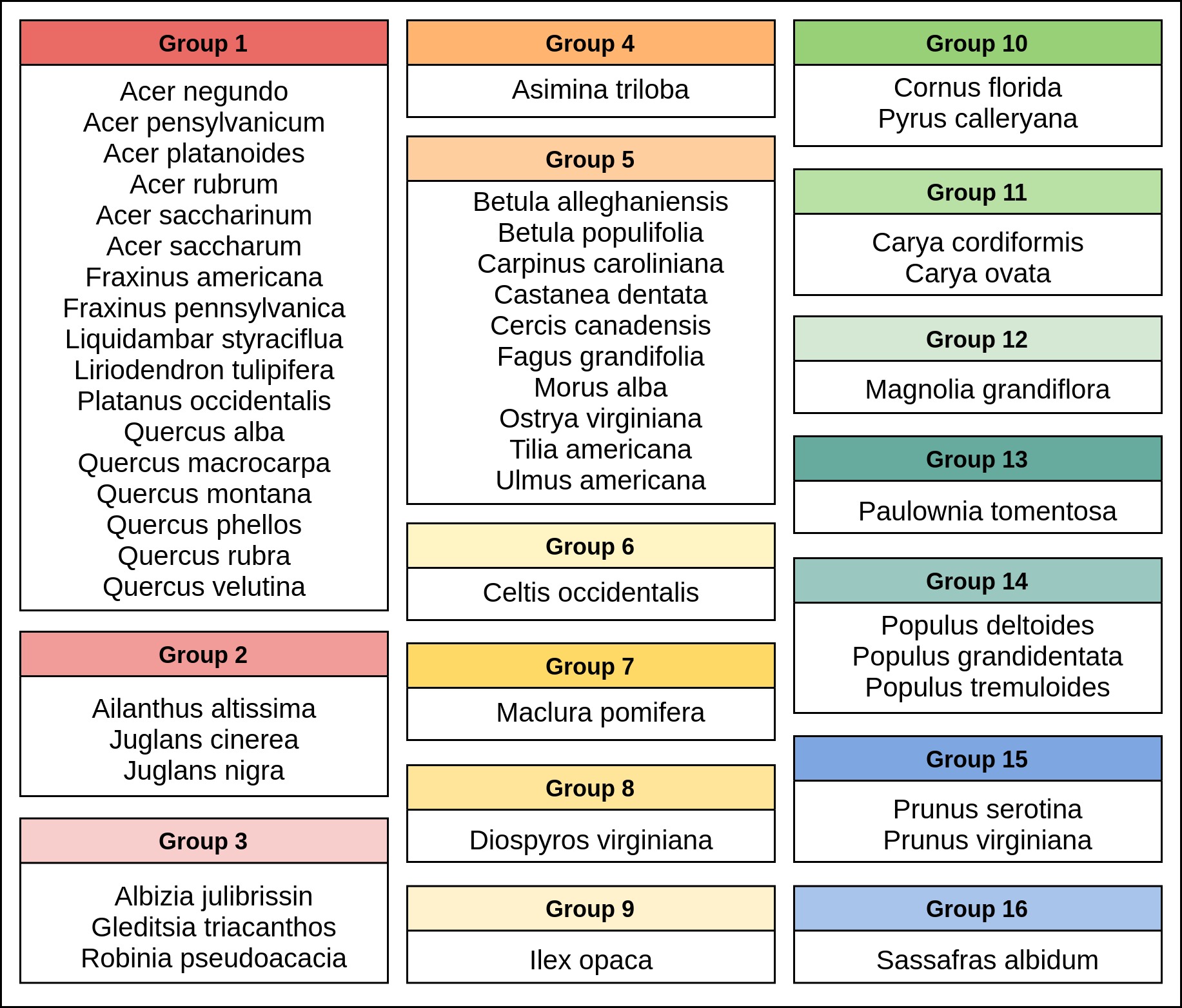}
			\caption{Top-50 similar tree species groups}
			\label{fig:speciesGroups}
		\end{figure}

		Further, iNaturalist provides data accompanying each species on the species it is most often confused with on the platform. We use the corresponding species data to create 16 similar species groups, listed in Figure~\ref{fig:speciesGroups}. Ideally, each group represents unique features visually characteristic of the 50 species dataset.
		
		In order to demonstrate species classification with minimal labelling effort, we selected 12 species (see Table~\ref{tab:labelledSpecies}) to annotate by considering tree features and the created similar species groups. From each of these species, 200 random images were sampled and annotated with bounding boxes using LabelImg \citep{labelImg}. Several human factors contribute to our natural image dataset being incredibly demanding: poor quality images, images containing more than one tree feature, and features from different species in a single image.
		
		\subsection{Deep Convolutional Neural Networks} \label{deepconv}
		The triumph of convolutional neural networks (CNNs) \citep{krizhevsky2017imagenet} has led to its widespread adoption in deep learning and continual advances in image classification. Initially, increased network depth \citep{simonyan2014very} produced an increase in CNN accuracy. This advance gave rise to important deep CNN architecture families, Inception networks \citep{szegedy2015going, szegedy2016rethinking, szegedy2017inception} and Residual networks (ResNets) \citep{he2016deep, he2016identity}, frequently used as benchmark models.
		
		Wide Residual Networks (WRNs) \citep{zagoruyko2016wide} further build on the residual connections of ResNets for improved classification accuracy. These networks introduce a widening factor that determines the number of convolutions per residual block. However, enormous neural networks can become clunky and computationally taxing.
		
		Therefore, MobileNets \citep{howard2017mobilenets, sandler2018mobilenetv2, howard2019searching} address the efficiency of CNNs. They utilise depthwise separable convolutions, inverted residual connections and neural architecture search for low latency models. Additionally, width and resolution multipliers adjust the optimal size of MobileNets. 
		
		Ultimately, EfficientNets \citep{tan2019efficientnet} consolidate these advances. This method builds on the efficient neural network architecture of MobileNets, while also presenting a compound coefficient for uniform scaling. By optimally scaling the network depth, width and resolution, EfficientNets achieve better efficiency and improve classification performance.
		
		\subsection{Deep semi-supervised learning}
		
		Given a labelled training dataset, supervised learning (SL), routinely used for image classification, aims to predict correct labels for previously unseen examples. Data augmentation is frequently employed for added regularisation and enhanced training data \citep{shorten2019survey}. Further, since prominent CNN models pre-trained on large datasets like ImageNet \citep{deng2009imagenet, russakovsky2015imagenet} are readily available, it has become common to employ SL in conjunction with transfer learning.	
		
		However, it seems that unconventional neural network applications of SL have been exhausted. In contrast to SL, semi-supervised learning (SSL) aims to utilise additional unlabelled training data for more accurate predictions \citep{oliver2018realistic}. The numerous different deep SSL methods largely follow one of two main approaches: methods employing consistency regularisation \citep{belkin2006manifold} and techniques that draw on entropy minimisation \citep{grandvalet2005semi}. 
		
		Methods based on consistency regularisation utilise unlabelled data through the idea that model predictions should be similar for augmented versions of the same image. Entropy minimisation techniques, like pseudo-labelling \citep{lee2013pseudo}, use the model to produce artificial labels for the unlabelled data.
		
		FixMatch \citep{sohn2020fixmatch}, with only an additional loss term compared to conventional SL, combines consistency regularisation and pseudo-labelling. This simple SSL method predicts artificial labels by using weakly-augmented unlabelled examples as the objective for strongly-augmented versions. Weak augmentation comprises random flip-and-shift augmentation, while strong augmentations are generated by, for instance, CTAugment \citep{berthelot2019remixmatch}.
		
		The consolidated FixMatch classification loss \citep{sohn2020fixmatch} being minimised for model parameters $\theta$ is
		\begin{equation}
			\ell_\mathrm{tot}(\theta) = \min_\theta (\ell_\mathrm{s}(\theta) + \lambda \ell_\mathrm{u}(\theta))
			\label{eqn:totalLossFixMatch}
		\end{equation}
		where $\ell_\mathrm{s}$ and $\ell_\mathrm{u}$ represent the supervised and unsupervised loss, respectively and $\lambda$ is the fixed scalar weight of the unsupervised loss. See \ref{FixMatchLosses} for a more complete description of the separate loss functions.
				
		\subsection{Method} \label{method}
		The compiled 50 species dataset comprises natural images with a collection of different tree features. Therefore, training an accurate deep learning classifier, which relies on visual similarities within a class, from the outset may prove difficult. In order to combat this, we propose a two-fold approach to tree species identification: tree feature recognition followed by species classification.
		
		Through feature recognition, we aim to distinguish leaves and bark from other tree features. Subsequently, we use this method to compile separate feature-specific datasets. Finally, we perform the classification of 50 species independently according to leaves and bark.
		
		\subsubsection{Tree feature recognition} \label{featureReg}
		 Our approach to feature recognition is binary classification with \textit{feature} and \textit{don't care} classes. The annotated 12 species dataset described in Section~\ref{dataset} includes 18 distinct tree features. We used this dataset to compile two feature datasets corresponding to leaves and bark.
		
\begin{figure}[ht!]
			\centering
			\begin{tabular}{ccc}
				\subfloat[simple leaf]{\includegraphics[width=1.5in]{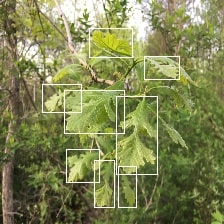} \label{fig:2refa}} &
				\subfloat[compounded leaf]{\includegraphics[width=1.5in]{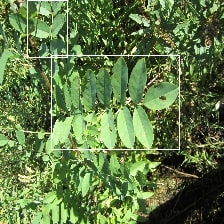} \label{fig:2refb}} &
				\subfloat[dry leaf]{\includegraphics[width=1.5in]{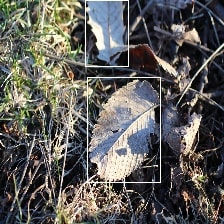} \label{fig:2refc}} \\
			\end{tabular}
			\caption{Feature examples with bounding box annotations for the leaves feature category. [Best viewed in colour]}
			\label{fig:annotatedExamplesLeaves}
		\end{figure}

\begin{figure}[ht!]
			\centering
			\begin{tabular}{cc}
				\subfloat[bark]{\includegraphics[width=1.5in]{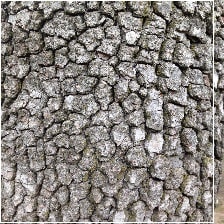} \label{fig:3refa}} &
				\subfloat[trunk]{\includegraphics[width=1.5in]{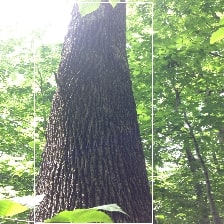} \label{fig:3refb}} \\
			\end{tabular}
			\caption{Feature examples with bounding box annotations for the bark feature category. [Best viewed in colour]}
			\label{fig:annotatedExamplesBark}
		\end{figure}

		Figures~\ref{fig:annotatedExamplesLeaves} and \ref{fig:annotatedExamplesBark} illustrate the different features belonging to each of the umbrella feature classes. For leaves, we include the simple leaf, compounded leaf and dry leaf features. Further, we consolidated images with bark and trunk annotations into the bark category. The don't care classes comprise the remainder of the images containing other features.	
		
		With more than 75\% of the species not included in the labelled feature datasets, a large set of 122 000 images are left unlabelled. Therefore, we employ the unlabelled data using our SSL method described in Section \ref{SSLmethod}.
		
		\subsubsection{Tree species classification} \label{speciesCls}		
		 We aim to perform the classification of 50 species according to differences in leaves and bark. Therefore, we use the tree feature recognition method introduced in Section~\ref{featureReg} to distinguish between different tree features. Feature predictions are made for the research-grade and need-ID datasets described in Section \ref{dataset}. Further, we group these images according to species to form discrete leaf and bark datasets.
		
\begin{figure}[ht!]
			\includegraphics[width=\linewidth]{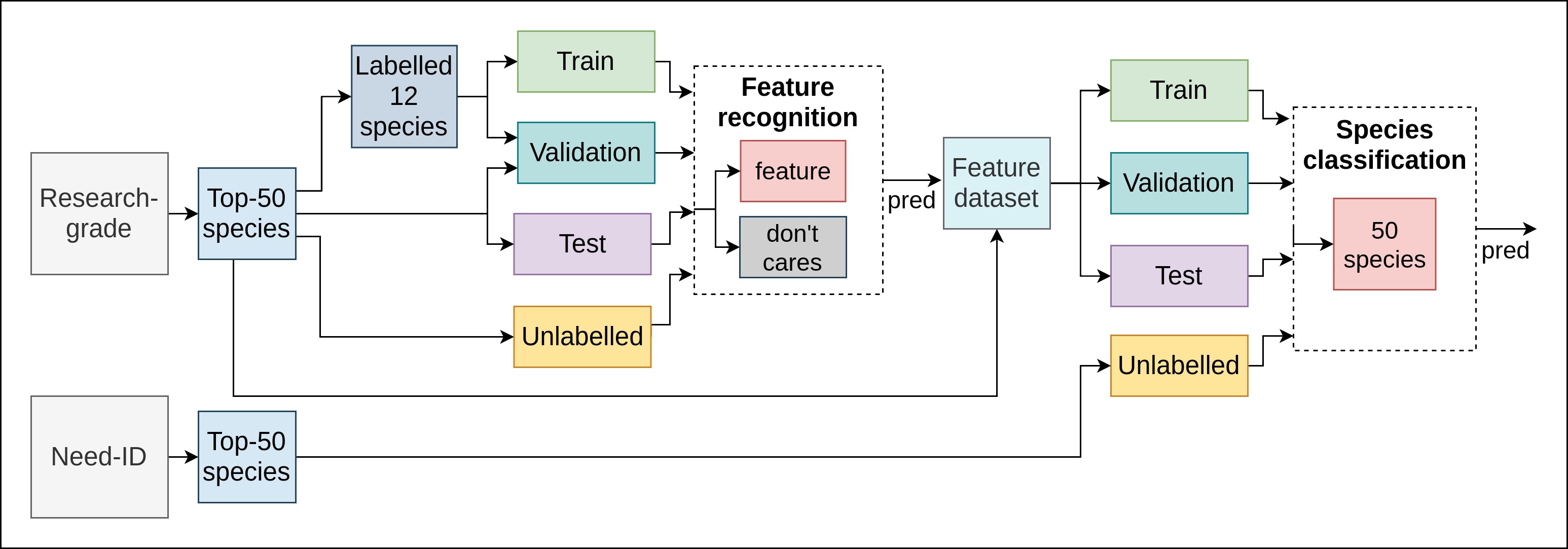}
			\caption{Method diagram for feature-specific species classification}
			\label{fig:methodDiagram}
		\end{figure}

		Since the need-ID dataset labels are not considered error-free and could adversely affect SL, we utilise this dataset in an unlabelled capacity. Consequently, species classification also employs SSL, using the same model as feature recognition for consistency. Figure~\ref{fig:methodDiagram} illustrates the complete workflow of our method.
		
		\subsubsection{Semi-supervised method} \label{SSLmethod}
		For our SSL method, we adopt FixMatch for its state-of-the-art yet straightforward approach. However, to further improve our implementation, we add enhancements to the vanilla FixMatch. 
		
			The Resnet-50 is a prevalent deep CNN benchmarking model. However, \citep{oliver2018realistic} proposed recommendations for SSL comparisons and adopted a WRN backbone model\footnote{deep neural network model updated during training}. Firstly, we exchange this WRN model, which the FixMatch implementation retained, for an EfficientNet model. Since it is memory efficient, we use EfficientNet-B0, the smallest EfficientNet.
		
		Further, according to the ratio $\mu$, let $\mathcal{X} = \{(x_b, p_b) : b \in (1, \dots, B)\}$ and $\mathcal{U} = \{u_b : b \in (1, \dots, \mu B)\}$ represent a batch of $B$ labelled and $\mu B$ unlabelled examples, respectively. We altered the supervised loss $\ell_\mathrm{s}$ (Appendix \ref{eqn:supervisedLoss}) to represent the cross-entropy on strongly-augmented examples $\mathcal{A}(x_b)$:
		\begin{equation}
			\ell_\mathrm{s} = \frac{1}{B}\sum^{B}_{b=1}H(p_b,p_\theta(y|\mathcal{A}(x_b))).
		\end{equation}
		
		The fixed value of the $\lambda$ weighting in Equation~\ref{eqn:totalLossFixMatch} is a known impediment of FixMatch performance through confirmation bias \citep{sohn2020fixmatch,arazo2020pseudo}. Fortunately, several recent techniques \citep{hendrycks2019augmix, rizve2021defense, ren2020not} address uncertainty estimation for SSL. We adopt the approach of \citep{ren2020not}, which aims to learn per-example unlabelled weights through an approximated influence function. For $\Lambda = \{\lambda_{u_b}: u_b \in \mathcal{U}\}$, Equation~\ref{eqn:totalLossFixMatch} can be reformulated as
		\begin{equation}
			\ell_\mathrm{tot}(\theta, \Lambda) = \min_\theta (\ell_\mathrm{s}(\theta) + \Lambda \cdot \ell_\mathrm{u}(\theta)).
			\label{eqn:totalLossNew}
		\end{equation}
		
		The unlabelled weights $\Lambda$, for model parameters $\theta^*$ which minimise $\ell_\mathrm{tot}$, are updated in \citep{ren2020not} through
		\begin{equation}
			\Lambda = \Lambda + \eta \cdot \bigtriangledown_\theta \ell_{\mathrm{v}}(\theta^*)^\top H_{\theta^*}^{-1} \bigtriangledown_\theta \ell_\mathrm{u}(\theta^*)
			\label{eqn:LambdaApprox}
		\end{equation}
		
		where $\eta$ represents the learning rate and $\bigtriangledown_\theta \ell_\mathrm{v}(\theta^*)$ indicates the gradient of the validation loss. Further, $H_{\theta^*} \triangleq \bigtriangledown_\theta^2 \ell_\mathrm{tot}(\theta^*)$ represents the Hessian and $\bigtriangledown_\theta \ell_\mathrm{u}(\theta^*)$ indicates the gradient of the per-example unlabelled loss. We refer the interested reader to \ref{FixMatchWeighting} for a more complete derivation of the influence function.
		
		To be consistent with \citep{sohn2020fixmatch}, we opt for the cosine learning rate decay \citep{loshchilov2016sgdr} for FixMatch. Similarly, for the unlabelled weighting this results in
		\begin{equation}
			\eta = \frac{5}{2}(1 + \cos (\frac{\pi k}{K}))
		\end{equation}
		
		with $k$ the current step and $K$ the total number of decay steps.
		
	\section{Results and Discussion} \label{Results}
	\subsection{Implementation details}
	 We adapt our SSL method from \citep{sohn2020fixmatch} and implement it in Tensorflow~2.3~\citep{tensorflow2015-whitepaper}. All methods employ EfficientNet-B0 as the backbone model with training image sizes of 224x224. We use the same code base and model architecture for all our method comparisons.
	
	 We conduct experiments on a single GPU with 11GB RAM, which limited the mini-batch sizes considerably compared to \citep{sohn2020fixmatch}. Feature recognition uses a mini-batch size of 6 and pseudo-label confidence threshold $\tau$ value of 0.98, while species classification uses mini-batches of 24 and $\tau$ of 0.8.
	
	We follow the original influence function implementation in \citep{ren2020not} for weighting the unlabelled data for feature recognition by using the penultimate layer gradients for computation. However, due to limited GPU memory, we had to alter the influence function in order to manage the increase in the number of classes for species classification. Therefore, we use the gradients of the final model layer for determining the influence function. We opted to update the influences every training iteration. 
	
	Further, to negate the effect of multiple features per image, we expand standard preprocessing techniques for all experiments using random cropping and distortion of the examples according to their bounding box annotations. We took care to train optimised SL and SSL models using the same CTAugment strategy. The reported results are that of the final checkpoint, and additional FixMatch and SL parameters related to our implementation are available in \ref{ImplementationParams}.
	\subsection{Tree feature recognition results}
	
	We evaluate our tree feature recognition method for leaves and bark separately. The test (480 examples) and validation set (720 examples) for each feature is randomly sampled from the 50 species dataset. Further, we compare our SSL method and its SL equivalent using an increasing number of labelled training data: 10\% (192 examples), 50\% (960 examples) and 100\% (1920 examples).
	
	In favour of coherence, we kept our implementation the same for both tree features. We train all models from scratch for 1000, 500 and 250 epochs for 10\%, 50\% and 100\% labelled data, respectively. The leaf and bark recognition results are presented in Table~\ref{tab:featureRecognitionResults}.
	
	\begin{table}[ht!]
		\centering
		\caption{Feature recognition accuracy for different dataset percentages using supervised (SL) and our semi-supervised method (SSL)}
		\begin{tabular}{ccccccc} 
			\toprule
			\multirow{2}{*}{\textbf{Feature}} & \multicolumn{2}{c}{\textbf{10\% labels}} & \multicolumn{2}{c}{\textbf{50\% labels}} & \multicolumn{2}{c}{\textbf{100\% labels}}	\\
									  & \textbf{SL}  & \textbf{SSL}  & \textbf{SL}  & \textbf{SSL}  & \textbf{SL}  & 	\textbf{SSL}        \\ \hline  
							Leaves    &    80.00	 & 	   83.13	 &	  86.67     &    87.50 	    & 	 86.87	   &    \textbf{87.71}			   \\    
							Bark      &    81.63	 & 	   83.09	 &	  86.43 	&    86.22	 	&\textbf{88.31}&	\textbf{88.31}			   \\	            \bottomrule
		\end{tabular}
		\label{tab:featureRecognitionResults}
	\end{table}
	
	Our leaf recognition SSL method outperforms SL on all the different dataset sizes. As would be expected, the performance gain is most significant with fewer labelled examples. Further, our method outperforms SL for bark recognition on the 10\% dataset and performs comparably on the remaining datasets.
	
	SSL scarcely employs transfer learning \citep{oliver2018realistic}. However, we pair our method with transfer learning using a stock Imagenet pre-trained EfficientNet-B0 model. We unfreeze the top 30 layers for training, and Table~\ref{tab:pretrainedFeatureRecognitionResults} shows the feature recognition results on the 100\% labelled feature datasets.
	
	\begin{table}[ht!]
		\centering
		\caption{Feature recognition accuracy for supervised (SL) and our semi-supervised method (SSL) using pre-trained EfficientNet-B0 models}
		\begin{tabular}{ccc} 
			\toprule
			\textbf{Feature} & \textbf{SL}  & \textbf{SSL}          \\ \hline  
				   Leaves    &   92.29 	 	& 	 \textbf{93.96}  	 				   \\    
				   Bark      &   90.61 		& 	 \textbf{93.11}  	 				   \\	            \bottomrule
		\end{tabular}
		\label{tab:pretrainedFeatureRecognitionResults}
	\end{table}
	
	Our SSL feature recognition method outperforms SL in utilising a pre-trained model for the full-label leaf and bark datasets. Table~\ref{tab:maskPercentages} shows the percentage increase of unlabelled example prediction confidence above the selected threshold $\tau$ of 0.98. This increase in prediction quality for unlabelled examples strengthens performance by increasing the number of unlabelled samples exploited during training.
	
	\begin{table}[ht!]
		\centering
		\caption{Unlabelled prediction confidence above the selected threshold $\tau$ of 0.98 for our SSL feature recognition method trained from scratch and using a pre-trained EfficientNet-B0 model}
		\begin{tabular}{ccc} 
			\toprule
			\textbf{Feature} & \textbf{SSL}  & \textbf{pre-trained SSL}          \\ \hline  
				   Leaves    &   50.46 	 	& 	 \textbf{84.42}  	 				   \\    
				   Bark      &   39.84 		& 	 \textbf{83.34}  	 				   \\	            \bottomrule
		\end{tabular}
		\label{tab:maskPercentages}
	\end{table}
	
\begin{figure}[ht!]
		\centering
		\setlength{\fboxsep}{0pt}
		\setlength{\fboxrule}{1pt}
		\begin{tabular}{cccc}	
			\subfloat[leaves]{\fcolorbox{green}{white}{\includegraphics[width=1.2in]{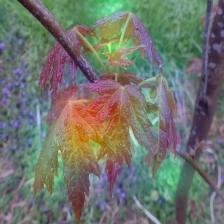}} \label{fig:5refa}} &
			\subfloat[leaves]{\fcolorbox{green}{white}{\includegraphics[width=1.2in]{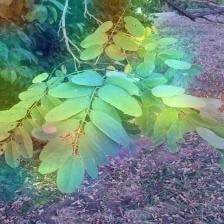}} \label{fig:5refb}} &
			\subfloat[leaves]{\fcolorbox{green}{white}{\includegraphics[width=1.2in]{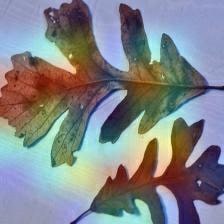}} \label{fig:5refc}} &
			\subfloat[don't care]{\fcolorbox{magenta}{white}{\includegraphics[width=1.2in]{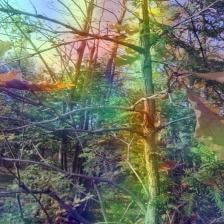}} \label{fig:5refd}} \\
		\end{tabular}
		\caption{Example predictions and Grad-CAM saliency maps for leaf recognition. Green and magenta borders are used to indicate correct and wrong predictions, respectively. [Best viewed in colour]}
		\label{fig:gradcamExamplesLeaves}
	\end{figure}

\begin{figure}[ht!]
		\centering
		\setlength{\fboxsep}{0pt}
		\setlength{\fboxrule}{1pt}
		\begin{tabular}{cccc}
			\subfloat[bark]{\fcolorbox{green}{white}{\includegraphics[width=1.2in]{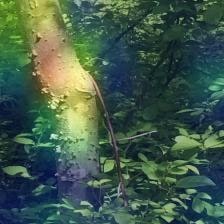}} \label{fig:6refa}} &
			\subfloat[bark]{\fcolorbox{green}{white}{\includegraphics[width=1.2in]{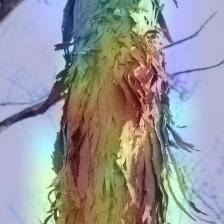}} \label{fig:6refb}} &
			\subfloat[bark]{\fcolorbox{green}{white}{\includegraphics[width=1.2in]{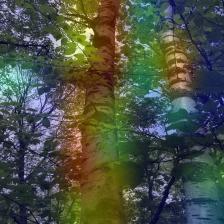}} \label{fig:6refc}} &
			\subfloat[don't care]{\fcolorbox{magenta}{white}{\includegraphics[width=1.2in]{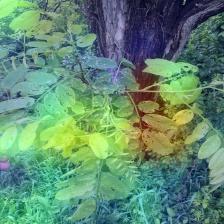}} \label{fig:6refd}} \\
		\end{tabular}
		\caption{Example predictions and Grad-CAM saliency maps for bark recognition. Green and magenta borders are used to indicate correct and wrong predictions, respectively. [Best viewed in colour]}
		\label{fig:gradcamExamplesBark}
	\end{figure}

	We use Grad-CAM \citep{selvaraju2017grad} for visual verification of our feature recognition method. Grad-CAM manipulates the gradient flowing into the final CNN layer to construct a saliency map that emphasises areas of interest in a classified image. Figures~\ref{fig:gradcamExamplesLeaves} and \ref{fig:gradcamExamplesBark} show Grad-CAM examples of our pre-trained SSL leaf and bark recognition model, respectively.
	
	An intuition into what our method prioritises for classification can be formed by looking at the highlighted areas. Further, the incorrect predictions illustrate the complexity of natural images by accentuating the don't care features favoured by the model.
	
	\subsection{Tree species classification results}
\begin{figure}[ht!]
		\centering
		\includegraphics[scale=0.075]{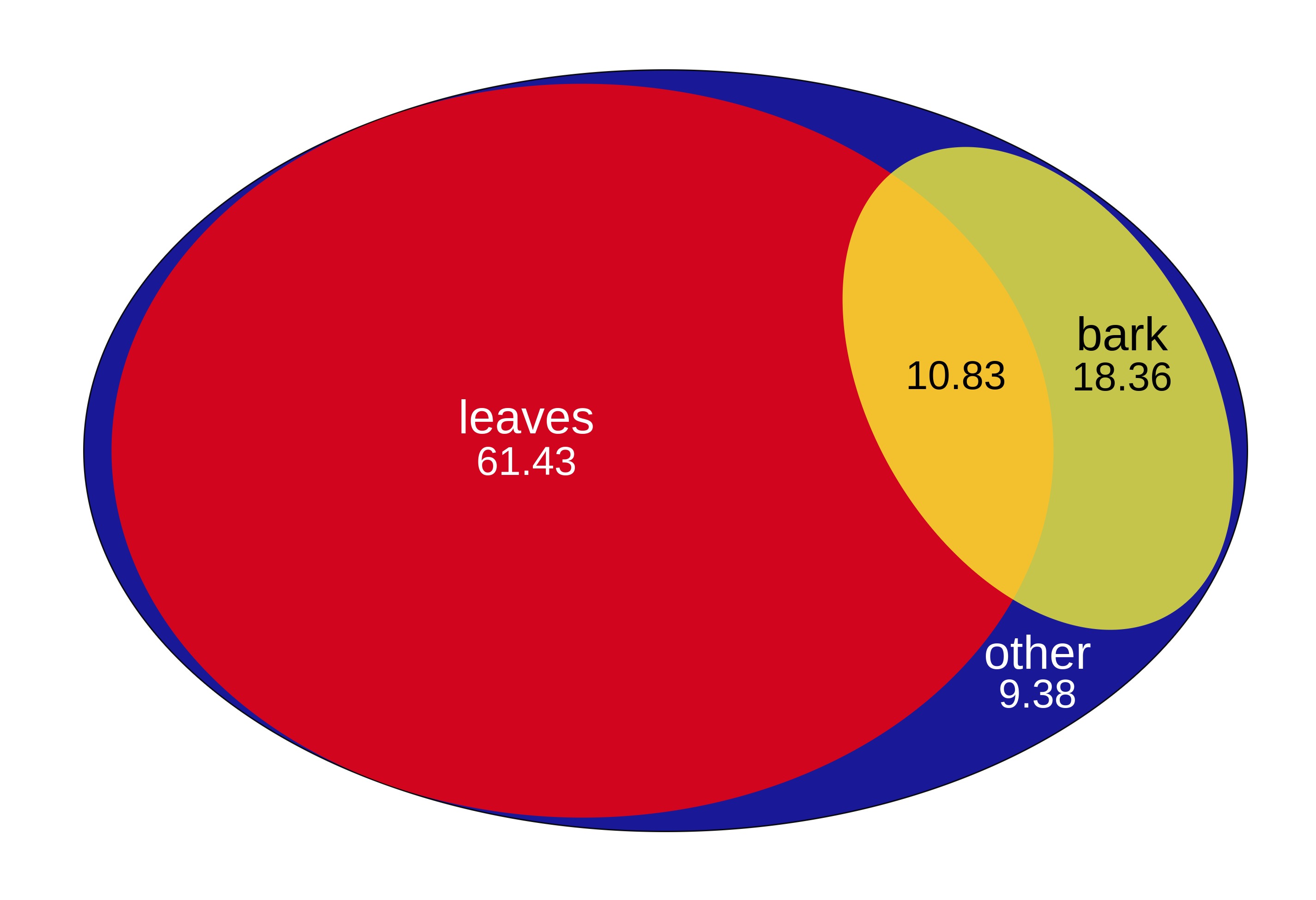}
		\caption{Feature prediction percentages for the 50 species research-grade dataset}
		\label{fig:featurePerc}
	\end{figure}

	We employ the leaf and bark feature recognition models to predict the features of the 50 species datasets. These predictions constitute separate feature-specific research-grade and need-ID datasets. Figure~\ref{fig:featurePerc} reinforces the complexity of the natural dataset by illustrating that the intersection of leaf and bark predictions is 10.83\% for the research-grade dataset.  
	
	Further, we group the feature-specific datasets according to species and use the corresponding research-grade dataset to generate the labelled training data. We compile a set-aside balanced test set of 20\%, and a validation set of 10\% labelled data, respectively. Our SSL method employs the need-ID feature datasets as unlabelled data. Table~\ref{tab:speciesClassificationDist} shows the resultant feature-specific dataset distributions used for species classification.
	
For training, we utilise the highly effective Imagenet pre-trained EfficientNet-B0 model used for feature recognition. We initialised the SSL unlabelled weights to 0 as it improved the stability of our weighting. Table~\ref{tab:top50SpeciesFeatures} shows the results of feature-specific species classification.
	
	\begin{table}[ht!]
		\centering
		\caption{Breakdown of feature-specific dataset distribution for species classification}
		\begin{tabular}{ccccccc}
			\toprule
			\textbf{Feature} & \textbf{Train}  & \textbf{Test} & \textbf{Validation}  & \textbf{Unlabelled}\\ \midrule
			Leaves	         &   63 854  	   &	18 287	   & 	9 156          	  &   29 044           \\ 
	    	Bark             &   25 770 	   &	7 399      &    3 528			  &   16 488           \\ \bottomrule
		\end{tabular}
		\label{tab:speciesClassificationDist}
	\end{table}
	
	\begin{table}[ht!]
		\centering
		\caption{Feature-specific 50 species classification accuracy using an Imagenet pre-trained EfficientNet-B0}
		\begin{tabular}{ccccccc}
			\toprule
			\multirow{2}{*}{\textbf{Method}} & \multicolumn{2}{c}{\textbf{Leaves}}  & \multicolumn{2}{c}{\textbf{Bark}} \\
					             & \textbf{Top-1} & \textbf{Top-5} & \textbf{Top-1} & \textbf{Top-5}  \\ \midrule
			SL	             	 & 78.12   		  &	93.94	 	   & 58.66	 		& 	82.66          \\ 
	    	SSL	              	 & \textbf{78.66} & \textbf{94.04} & \textbf{59.91}	&  \textbf{83.04}        \\ \bottomrule
		\end{tabular}
		\label{tab:top50SpeciesFeatures}
	\end{table}

	 Our novel SSL method utilises in-domain unlabelled examples to outperform SL for species classification according to leaves. Further, our method also obtains superior results for bark species classification.
	
	\section{Conclusions} \label{Conclusions}
	We compile and present natural tree image datasets for species classification. Our novel two-fold approach includes feature recognition and species classification. Further, with limited training data, we utilise SSL for our implementation.
	
	We introduce an SSL-based tree feature recognition method for natural images which supports leaves and bark. Our method outperforms standard SL for leaf recognition and performs comparably to SL for bark recognition of 50 species using limited annotated data. Although seldomly paired with SSL, we employ pre-trained models in conjunction with SSL for feature recognition. We demonstrate superior feature recognition to pre-trained SL with our best models achieving 93.96\% and 93.11\% accuracy for leaves and bark, respectively.
	
	In addition, we show an application of our feature recognition method by generating labelled and unlabelled feature-specific datasets for bark and leaves. We use these datasets to illustrate further the versatility and performance of our method for species classification. Our fine-tuned SSL implementation outperforms SL for species classification according to leaf and bark with 94.04\% and 83.04\% top-5 accuracy, respectively.
	
	\section*{Acknowledgements}
		We gratefully acknowledge the support of NVIDIA Corporation with the donation of the Titan X Pascal GPU used for this research.
	
\newpage
\bibliographystyle{elsarticle-num}
\bibliography{tree_species_classification_references}

\begin{thebibliography}{10}
\expandafter\ifx\csname url\endcsname\relax
  \def\url#1{\texttt{#1}}\fi
\expandafter\ifx\csname urlprefix\endcsname\relax\def\urlprefix{URL }\fi
\expandafter\ifx\csname href\endcsname\relax
  \def\href#1#2{#2} \def\path#1{#1}\fi

\bibitem{pimm2014biodiversity}
S.~L. Pimm, C.~N. Jenkins, R.~Abell, T.~M. Brooks, J.~L. Gittleman, L.~N.
  Joppa, P.~H. Raven, C.~M. Roberts, J.~O. Sexton, The biodiversity of species
  and their rates of extinction, distribution, and protection, science
  344~(6187).

\bibitem{waldchen2018automated}
J.~W{\"a}ldchen, M.~Rzanny, M.~Seeland, P.~M{\"a}der, Automated plant species
  identification - trends and future directions, PLoS computational biology
  14~(4) (2018) e1005993.

\bibitem{martin2001database}
D.~Martin, C.~Fowlkes, D.~Tal, J.~Malik, A database of human segmented natural
  images and its application to evaluating segmentation algorithms and
  measuring ecological statistics, in: Proceedings Eighth IEEE International
  Conference on Computer Vision. ICCV 2001, Vol.~2, IEEE, 2001, pp. 416--423.

\bibitem{inaturalist}
{iNaturalist}, \url{https://www.inaturalist.org/}, [Online; accessed:
  23-04-2019].

\bibitem{Horn_2018_CVPR}
G.~Van~Horn, O.~Mac~Aodha, Y.~Song, Y.~Cui, C.~Sun, A.~Shepard, H.~Adam,
  P.~Perona, S.~Belongie, The inaturalist species classification and detection
  dataset, in: Proceedings of the IEEE Conference on Computer Vision and
  Pattern Recognition (CVPR), 2018.

\bibitem{carpentier2018tree}
M.~Carpentier, P.~Gigu{\`e}re, J.~Gaudreault, Tree species identification from
  bark images using convolutional neural networks, in: 2018 IEEE/RSJ
  International Conference on Intelligent Robots and Systems (IROS), IEEE,
  2018, pp. 1075--1081.

\bibitem{wu2007leaf}
S.~G. Wu, F.~S. Bao, E.~Y. Xu, Y.-X. Wang, Y.-F. Chang, Q.-L. Xiang, A leaf
  recognition algorithm for plant classification using probabilistic neural
  network, in: 2007 IEEE international symposium on signal processing and
  information technology, IEEE, 2007, pp. 11--16.

\bibitem{simpson1992biological}
R.~Simpson, R.~Williams, R.~Ellis, P.~Culverhouse, Biological pattern
  recognition by neural networks, Marine Ecology Progress Series (1992)
  303--308.

\bibitem{morris1992identification}
C.~Morris, L.~Boddy, R.~Allman, Identification of basidiomycete spores by
  neural network analysis of flow cytometry data, Mycological Research 96~(8)
  (1992) 697--701.

\bibitem{pawara2017comparing}
P.~Pawara, E.~Okafor, O.~Surinta, L.~Schomaker, M.~Wiering, Comparing local
  descriptors and bags of visual words to deep convolutional neural networks
  for plant recognition, in: International Conference on Pattern Recognition
  Applications and Methods, Vol.~2, SCITEPRESS, 2017, pp. 479--486.

\bibitem{vsulc2017fine}
M.~{\v{S}}ulc, J.~Matas, Fine-grained recognition of plants from images, Plant
  Methods 13~(1) (2017) 115.

\bibitem{bourlard1996towards}
H.~Bourlard, H.~Hermansky, N.~Morgan, Towards increasing speech recognition
  error rates, Speech communication 18~(3) (1996) 205--231.

\bibitem{pise2008survey}
N.~N. Pise, P.~Kulkarni, A survey of semi-supervised learning methods, in: 2008
  International Conference on Computational Intelligence and Security, Vol.~2,
  IEEE, 2008, pp. 30--34.

\bibitem{dalponte2015semi}
M.~Dalponte, L.~T. Ene, M.~Marconcini, T.~Gobakken, E.~N{\ae}sset,
  Semi-supervised svm for individual tree crown species classification, ISPRS
  Journal of Photogrammetry and Remote Sensing 110 (2015) 77--87.

\bibitem{weinstein2019individual}
B.~G. Weinstein, S.~Marconi, S.~Bohlman, A.~Zare, E.~White, Individual
  tree-crown detection in rgb imagery using semi-supervised deep learning
  neural networks, Remote Sensing 11~(11) (2019) 1309.

\bibitem{kumar2012leafsnap}
N.~Kumar, P.~N. Belhumeur, A.~Biswas, D.~W. Jacobs, W.~J. Kress, I.~C. Lopez,
  J.~V. Soares, Leafsnap: A computer vision system for automatic plant species
  identification, in: European conference on computer vision, Springer, 2012,
  pp. 502--516.

\bibitem{labelImg}
Tzutalin, Labelimg, \url{https://github.com/tzutalin/labelImg}, [Git code]
  (2015).

\bibitem{krizhevsky2017imagenet}
A.~Krizhevsky, I.~Sutskever, G.~E. Hinton, Imagenet classification with deep
  convolutional neural networks, Communications of the ACM 60~(6) (2017)
  84--90.

\bibitem{simonyan2014very}
K.~Simonyan, A.~Zisserman, Very deep convolutional networks for large-scale
  image recognition, arXiv preprint arXiv:1409.1556.

\bibitem{szegedy2015going}
C.~Szegedy, W.~Liu, Y.~Jia, P.~Sermanet, S.~Reed, D.~Anguelov, D.~Erhan,
  V.~Vanhoucke, A.~Rabinovich, Going deeper with convolutions, in: Proceedings
  of the IEEE conference on computer vision and pattern recognition, 2015, pp.
  1--9.

\bibitem{szegedy2016rethinking}
C.~Szegedy, V.~Vanhoucke, S.~Ioffe, J.~Shlens, Z.~Wojna, Rethinking the
  inception architecture for computer vision, in: Proceedings of the IEEE
  conference on computer vision and pattern recognition, 2016, pp. 2818--2826.

\bibitem{szegedy2017inception}
C.~Szegedy, S.~Ioffe, V.~Vanhoucke, A.~Alemi, Inception-v4, inception-resnet
  and the impact of residual connections on learning, in: Proceedings of the
  AAAI Conference on Artificial Intelligence, Vol.~31, 2017.

\bibitem{he2016deep}
K.~He, X.~Zhang, S.~Ren, J.~Sun, Deep residual learning for image recognition,
  in: Proceedings of the IEEE conference on computer vision and pattern
  recognition, 2016, pp. 770--778.

\bibitem{he2016identity}
K.~He, X.~Zhang, S.~Ren, J.~Sun, Identity mappings in deep residual networks,
  in: European conference on computer vision, Springer, 2016, pp. 630--645.

\bibitem{zagoruyko2016wide}
S.~Zagoruyko, N.~Komodakis, Wide residual networks, arXiv preprint
  arXiv:1605.07146.

\bibitem{howard2017mobilenets}
A.~G. Howard, M.~Zhu, B.~Chen, D.~Kalenichenko, W.~Wang, T.~Weyand,
  M.~Andreetto, H.~Adam, Mobilenets: Efficient convolutional neural networks
  for mobile vision applications, arXiv preprint arXiv:1704.04861.

\bibitem{sandler2018mobilenetv2}
M.~Sandler, A.~Howard, M.~Zhu, A.~Zhmoginov, L.-C. Chen, Mobilenetv2: Inverted
  residuals and linear bottlenecks, in: Proceedings of the IEEE conference on
  computer vision and pattern recognition, 2018, pp. 4510--4520.

\bibitem{howard2019searching}
A.~Howard, M.~Sandler, G.~Chu, L.-C. Chen, B.~Chen, M.~Tan, W.~Wang, Y.~Zhu,
  R.~Pang, V.~Vasudevan, et~al., Searching for mobilenetv3, in: Proceedings of
  the IEEE International Conference on Computer Vision, 2019, pp. 1314--1324.

\bibitem{tan2019efficientnet}
M.~Tan, Q.~V. Le, Efficientnet: Rethinking model scaling for convolutional
  neural networks, arXiv preprint arXiv:1905.11946.

\bibitem{shorten2019survey}
C.~Shorten, T.~M. Khoshgoftaar, A survey on image data augmentation for deep
  learning, Journal of Big Data 6~(1) (2019) 60.

\bibitem{deng2009imagenet}
J.~Deng, W.~Dong, R.~Socher, L.-J. Li, K.~Li, L.~Fei-Fei, Imagenet: A
  large-scale hierarchical image database, in: 2009 IEEE conference on computer
  vision and pattern recognition, Ieee, 2009, pp. 248--255.

\bibitem{russakovsky2015imagenet}
O.~Russakovsky, J.~Deng, H.~Su, J.~Krause, S.~Satheesh, S.~Ma, Z.~Huang,
  A.~Karpathy, A.~Khosla, M.~Bernstein, et~al., Imagenet large scale visual
  recognition challenge, International journal of computer vision 115~(3)
  (2015) 211--252.

\bibitem{oliver2018realistic}
A.~Oliver, A.~Odena, C.~A. Raffel, E.~D. Cubuk, I.~Goodfellow, Realistic
  evaluation of deep semi-supervised learning algorithms, Advances in neural
  information processing systems 31 (2018) 3235--3246.

\bibitem{belkin2006manifold}
M.~Belkin, P.~Niyogi, V.~Sindhwani, Manifold regularization: A geometric
  framework for learning from labeled and unlabeled examples., Journal of
  machine learning research 7~(11).

\bibitem{grandvalet2005semi}
Y.~Grandvalet, Y.~Bengio, et~al., Semi-supervised learning by entropy
  minimization., CAP 367 (2005) 281--296.

\bibitem{lee2013pseudo}
D.-H. Lee, et~al., Pseudo-label: The simple and efficient semi-supervised
  learning method for deep neural networks, in: Workshop on challenges in
  representation learning, ICML, Vol.~3, 2013, p. 896.

\bibitem{sohn2020fixmatch}
K.~Sohn, D.~Berthelot, C.-L. Li, Z.~Zhang, N.~Carlini, E.~D. Cubuk, A.~Kurakin,
  H.~Zhang, C.~Raffel, Fixmatch: Simplifying semi-supervised learning with
  consistency and confidence, arXiv preprint arXiv:2001.07685.

\bibitem{berthelot2019remixmatch}
D.~Berthelot, N.~Carlini, E.~D. Cubuk, A.~Kurakin, K.~Sohn, H.~Zhang,
  C.~Raffel, Remixmatch: Semi-supervised learning with distribution matching
  and augmentation anchoring, in: International Conference on Learning
  Representations, 2019.

\bibitem{arazo2020pseudo}
E.~Arazo, D.~Ortego, P.~Albert, N.~E. O’Connor, K.~McGuinness,
  Pseudo-labeling and confirmation bias in deep semi-supervised learning, in:
  2020 International Joint Conference on Neural Networks (IJCNN), IEEE, 2020,
  pp. 1--8.

\bibitem{hendrycks2019augmix}
D.~Hendrycks, N.~Mu, E.~D. Cubuk, B.~Zoph, J.~Gilmer, B.~Lakshminarayanan,
  Augmix: A simple data processing method to improve robustness and
  uncertainty, arXiv preprint arXiv:1912.02781.

\bibitem{rizve2021defense}
M.~N. Rizve, K.~Duarte, Y.~S. Rawat, M.~Shah, In defense of pseudo-labeling: An
  uncertainty-aware pseudo-label selection framework for semi-supervised
  learning, arXiv preprint arXiv:2101.06329.

\bibitem{ren2020not}
Z.~Ren, R.~Yeh, A.~Schwing, Not all unlabeled data are equal: Learning to
  weight data in semi-supervised learning, Advances in Neural Information
  Processing Systems 33.

\bibitem{loshchilov2016sgdr}
I.~Loshchilov, F.~Hutter, Sgdr: Stochastic gradient descent with warm restarts,
  arXiv preprint arXiv:1608.03983.

\bibitem{tensorflow2015-whitepaper}
M.~Abadi, A.~Agarwal, P.~Barham, E.~Brevdo, Z.~Chen, C.~Citro, G.~S. Corrado,
  A.~Davis, J.~Dean, M.~Devin, S.~Ghemawat, I.~Goodfellow, A.~Harp, G.~Irving,
  M.~Isard, Y.~Jia, R.~Jozefowicz, L.~Kaiser, M.~Kudlur, J.~Levenberg,
  D.~Man\'{e}, R.~Monga, S.~Moore, D.~Murray, C.~Olah, M.~Schuster, J.~Shlens,
  B.~Steiner, I.~Sutskever, K.~Talwar, P.~Tucker, V.~Vanhoucke, V.~Vasudevan,
  F.~Vi\'{e}gas, O.~Vinyals, P.~Warden, M.~Wattenberg, M.~Wicke, Y.~Yu,
  X.~Zheng, \href{https://www.tensorflow.org/}{{TensorFlow}: Large-scale
  machine learning on heterogeneous systems}, software available from
  tensorflow.org (2015).
\newline\urlprefix\url{https://www.tensorflow.org/}

\bibitem{selvaraju2017grad}
R.~R. Selvaraju, M.~Cogswell, A.~Das, R.~Vedantam, D.~Parikh, D.~Batra,
  Grad-cam: Visual explanations from deep networks via gradient-based
  localization, in: Proceedings of the IEEE international conference on
  computer vision, 2017, pp. 618--626.

\end{thebibliography}

\newpage
\appendix
	\section{Feature Recognition}
	\begin{table}[ht]
		\caption{Number of annotated training images for the feature recognition datasets. Colours correspond to the similar species groups.}
		\resizebox{\textwidth}{!}{%
		\begin{tabular}{ccc|cc}
		\hline
		\textbf{Species} & \textbf{Leaves} & \textbf{Don't cares} & \textbf{Bark} & \textbf{Don't cares}       		 			\\ \hline
		\cellcolor[HTML]{ea6b66} Acer saccharinum                 & 107             & 59           	& 60   			& 110       \\
		\cellcolor[HTML]{b9e0a5} Carya ovata                      & 31              & 132          	& 121   		& 38      	\\
		\cellcolor[HTML]{fff4c3} Celtis occidentalis              & 81              & 80          	& 104   		& 61      	\\
		\cellcolor[HTML]{97d077} Cornus florida                   & 119             & 44          	& 34    		& 124       \\
		\cellcolor[HTML]{ea6b66} Fraxinus americana               & 62              & 86          	& 104    		& 46      	\\
		\cellcolor[HTML]{f8cecc} Gleditsia triacanthos            & 34              & 124          	& 108    		& 56      	\\
		\cellcolor[HTML]{f19c99} Juglans nigra                    & 103             & 53         	& 32   			& 132     	\\
		\cellcolor[HTML]{ffce9f} Morus alba                       & 150             & 3           	& 17   			& 150      	\\
		\cellcolor[HTML]{9ac7bf} Populus tremuloides              & 113             & 49          	& 62   			& 93      	\\
		\cellcolor[HTML]{ea6b66} Quercus macrocarpa               & 97              & 71          	& 73   			& 87      	\\
		\cellcolor[HTML]{f8cecc} Robinia pseudoacacia             & 110             & 52          	& 40   			& 113      	\\
		\cellcolor[HTML]{ffce9f} Ulmus americana                  & 59              & 101          	& 66   			& 89      	\\ \hline
														  & \textbf{1 066}	& \textbf{854}	& \textbf{821}	& \textbf{1 099} 	\\ \hline
\end{tabular}%
}
\label{tab:labelledSpecies}
\end{table}

	\section{FixMatch Loss} \label{FixMatchLosses}
	The FixMatch losses, as described in \citep{sohn2020fixmatch}, are included here for completeness. Given that the standard cross-entropy between probability distributions $p$ and $q$ is denoted by $H(p,q)$ and the predicted class distribution for input $x$ is $p_\theta(y|x)$.
	
	The supervised loss on batches of weakly-augmented labelled training examples $\alpha(x_b)$ is described by
		\begin{equation}
			\ell_\mathrm{s} = \frac{1}{B}\sum^{B}_{b=1}H(p_b,p_\theta(y|\alpha(x_b)))
			\label{eqn:supervisedLoss}
		\end{equation} 
		with $p_b$ representing the one-hot label distribution.
		
		Further, the prediction of an artificial label for the unsupervised loss uses a weakly-augmented unlabelled example prediction $q_b = p_\theta(y|\alpha(u_b))$ and enforces it against a strongly-augmented version $\mathcal{A}(u_b)$:
		\begin{equation}
			\ell_\mathrm{u} = \frac{1}{\mu B}\sum^{\mu B}_{b=1} \mathbb{1} (\textup{max}(q_b)\geq\tau)H(\hat{q_b},p_\theta(y|\mathcal{A}(u_b)))
		\end{equation}
		where $\hat{q_b} = \textup{argmax}(q_b)$ is the predicted pseudo-label and $\tau$ indicates the pseudo-label confidence threshold value. 
		
	\section{FixMatch Weighting} \label{FixMatchWeighting}
		The bi-level optimization problem in \citep{ren2020not} can be represented as
		
		\begin{equation}
			\min_{\Lambda} \ell_\mathrm{v}(\theta^*(\Lambda)) \ni \theta^*(\Lambda) = \arg \ell_\mathrm{tot}(\theta, \Lambda)
			\label{eqn:weightsOptimization}
		\end{equation}
		
		with validation loss
		\begin{equation}
			\ell_\mathrm{v} = \frac{1}{|\Lambda|}\sum^{|\Lambda|}_{b=1}H(p_b, p_m(y|\alpha(v_b)))
		\end{equation}
		
		for weakly-augmented labelled validation example batches $\alpha(v_b)$.
		
		Updating the per-example weights $\Lambda$ using Stochastic Gradient Descent (SGD) according to Equation~\ref{eqn:weightsOptimization} results in
		\begin{equation}
			\Lambda = \Lambda - \eta \cdot \frac{\partial \ell_\mathrm{v}(\theta^*(\Lambda))}{\partial\Lambda}
		\end{equation}
		which, since optimizing $\theta^*$ depends on $\Lambda$, results in Equation~\ref{eqn:LambdaApprox}.
		
	\section{Media sources used in this paper}
	\begin{itemize}
		\item Figure \ref{fig:2refb}: derivative of Photo 5850954 by Winooski Valley Park District via iNaturalist, used under \href{https://creativecommons.org/licenses/by-nc/4.0/}{CC BY-NC 4.0}
		\item Figure \ref{fig:2refc}: derivative of Photo 33826578 by andrewhipp via iNaturalist, used under \href{https://creativecommons.org/licenses/by-nc/4.0/}{CC BY-NC 4.0}
		\item Figure \ref{fig:3refa}: derivative of Photo 6923646 by Emily VandenBerg via iNaturalist, used under \href{https://creativecommons.org/licenses/by-nc/4.0/}{CC BY-NC 4.0}
		\item Figure \ref{fig:3refb}: derivative of Photo 990092 by Vermont Natural Heritage Inventory via iNaturalist, used under \href{https://creativecommons.org/licenses/by-nc/4.0/}{CC BY-NC 4.0}
		\item Figure \ref{fig:5refa}: derivative of Photo 16397116 by botanygirl via iNaturalist, used under \href{https://creativecommons.org/licenses/by/4.0/}{CC BY 4.0}
		\item Figure \ref{fig:5refb}: derivative of Photo 5408414 by elizm480 via iNaturalist, used under \href{https://creativecommons.org/licenses/by-nc/4.0/}{CC BY-NC 4.0}
		\item Figure \ref{fig:5refc}: derivative of Photo 16071078 by hobiecat via iNaturalist, used under \href{https://creativecommons.org/licenses/by-nc/4.0/}{CC BY-NC 4.0}
		\item Figure \ref{fig:5refd}: derivative of Photo 14382322 by Owen Clarkin via iNaturalist, used under \href{https://creativecommons.org/licenses/by-nc/4.0/}{CC BY-NC 4.0}
		\item Figure \ref{fig:6refa}: derivative of Photo 7726610 by jonibaum via iNaturalist, used under \href{https://creativecommons.org/licenses/by-nc/4.0/}{CC BY-NC 4.0}
		\item Figure \ref{fig:6refb}: derivative of Photo 16425992, \copyright Martin J. Calabrese, used with permission
		\item Figure \ref{fig:6refc}: derivative of Photo 9954956 by leojpa, used under \href{https://creativecommons.org/licenses/by-nc/4.0/}{CC BY-NC 4.0}
		\item Figure \ref{fig:6refd}: derivative of Photo 7655524 by John Boback via iNaturalist, used under \href{https://creativecommons.org/licenses/by-nc/4.0/}{CC BY-NC 4.0}
	\end{itemize}
	
	\section{Implementation Parameters} \label{ImplementationParams}
	The FixMatch parameters used for our implementation are set out in Table~\ref{tab:FixMatchSettings} below. Small batch sizes are due to limited available GPU memory. Further, the SL training parameters are listed in Table~\ref{tab:SLSettings}.
	
	\begin{table}[ht]
		\centering
		\caption{FixMatch parameter values}
		\begin{tabular}{ccc}
			\toprule
			\textbf{Variable} & \textbf{Feature Recognition} & \textbf{Species Classification}   							   \\ \midrule
			$\tau$							  &  0.98          	 									& 0.85    				   \\ 
			$\mu$		      				  &  7      		 									& 2      				   \\ 
			$B$				  				  &  6				 									& 24     				   \\
			wd			  				  	  &  0.0001			 									& 0.0001 				   \\
			$\mathcal{A}$					  &  CTA \citep{sohn2020fixmatch} (default parameters) 	& CTA (default parameters) \\					
	    	$\Lambda$	      				  &  $0.5 \in [0,2]$ 									& $0.0 \in [0,2]$          \\ 
	    	optimiser						  &  SGD 												& SGD					   \\
	    	learning rate					  &  Cosine decay										& Cosine decay			   \\
	    	base lr							  &  0.1												& 0.001					   \\ \bottomrule
		\end{tabular}
		\label{tab:FixMatchSettings}
	\end{table}
	
	\begin{table}[ht]
		\centering
		\caption{SL parameter values}
		\begin{tabular}{ccc}
			\toprule
			\textbf{Variable} & \textbf{Feature Recognition} & \textbf{Species Classification}   							   \\ \midrule
			$B$				  				  &  6				 									& 16     				   \\
			wd			  				  	  &  0.0001			 									& 0.0001 				   \\
			$\mathcal{A}$					  &  CTA (default parameters) 							& CTA (default parameters) \\		
			optimiser						  &  SGD 												& SGD					   \\
	    	learning rate					  &  Cosine decay										& Cosine decay			   \\	
    		base lr							  &  $390.625\times10^{-6}$								& $390.625\times10^{-6}$   \\ \bottomrule	  
		\end{tabular}
		\label{tab:SLSettings}
	\end{table}

\end{document}